%% file: naacl2022.tex
\documentclass[11pt,a4paper]{article}
\usepackage[final]{acl}
\usepackage{times}
\usepackage{latexsym}
\usepackage{siunitx}

\usepackage{microtype}

\input{abbrevs}

\usepackage{inconsolata}

\title{Same Neurons, Different Languages: \\ Probing Morphosyntax in Multilingual Pre-trained Models}

\author{
Karolina Stańczak$^{1}$ \quad
Edoardo Ponti$^{2}$ \quad
Lucas Torroba Hennigen$^{3}$ \\
\textbf{Ryan Cotterell}$^{4}$ \quad
\textbf{Isabelle Augenstein}$^{1}$ \\
$^1$University of Copenhagen \quad
$^2$Mila/McGill University\\
$^3$Massachusetts Institute of Technology \quad $^4$ETH Z{\"u}rich\\
{\tt \href{mailto:ks@di.ku.dk}{ks@di.ku.dk}} \quad {\tt \href{mailto:augenstein@di.ku.dk}{augenstein@di.ku.dk}} \quad {\tt \href{mailto:edoardo-maria.ponti@mila.quebec}{edoardo-maria.ponti@mila.quebec}} \\ 
{\tt \href{mailto:lucastor@mit.edu}{lucastor@mit.edu}} \quad {\tt \href{mailto:rcotterell@inf.ethz.ch}{rcotterell@inf.ethz.ch}}
}

\date{}

\begin{document}
\maketitle
\begin{abstract}
The success of multilingual pre-trained models
is underpinned by their ability to learn representations shared by multiple languages even in absence of any explicit supervision. 
However, it remains unclear \textit{how} these models learn to generalise across languages. 
In this work, we conjecture that multilingual pre-trained models can derive language-universal abstractions about grammar.
In particular, we investigate whether morphosyntactic information is encoded in the same subset of neurons in different languages.
We conduct the first large-scale empirical study over \XX 
languages and 14 morphosyntactic categories with a state-of-the-art neuron-level probe.
Our findings show that the cross-lingual overlap between neurons is significant, but its extent may vary across categories and depends on language proximity and pre-training data size.
\newline
\newline
\vspace{1.5em}
\hspace{.5em}\includegraphics[width=1.25em,height=1.25em]{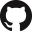}\hspace{.75em}\parbox{\dimexpr\linewidth-2\fboxsep-2\fboxrule}{\url{https://github.com/copenlu/multilingual-typology-probing}}
\vspace{-.5em}
\end{abstract}

\section{Introduction}
\label{sec:intro}
\edo{I did a pass throughout the Introduction, trying to adress all the outstanding comments.}
Massively multilingual pre-trained models \citep[\textit{inter alia}]{devlin-etal-2019-bert,conneau-etal-2020-unsupervised,liu-etal-2020-multilingual-denoising,xue-etal-2021-mt5} display an impressive ability to transfer knowledge between languages as well as to perform zero-shot learning \citep[\textit{inter alia}]{pires-etal-2019-multilingual,wu-dredze-2019-beto,nooralahzadeh-etal-2020-zero,hardalov2021fewshot}. 
Nevertheless, it remains unclear how pre-trained models actually manage to learn multilingual representations \emph{despite} the lack of an explicit signal through parallel texts. 
Hitherto, many have speculated that the overlap of sub-words between cognates in related languages plays a key role in the process of multilingual generalisation 
\citep{wu-dredze-2019-beto,Cao2020Multilingual,pires-etal-2019-multilingual,abendLexicalEventOrdering2015,vulic-etal-2020-probing}.

\begin{figure}[t]
    \centering
    \includegraphics[width=\linewidth]{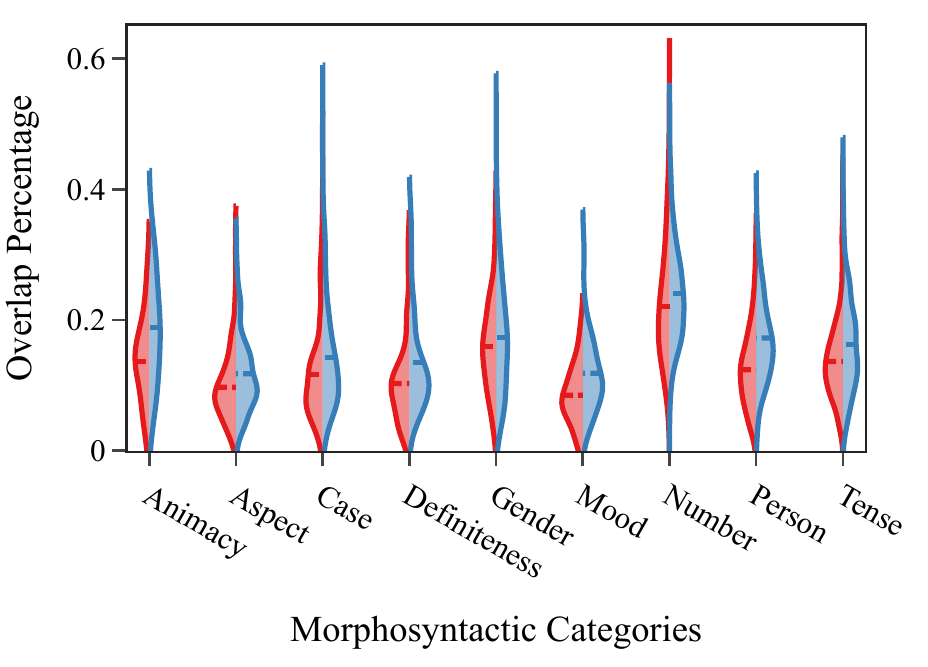}
    \caption{Percentages of neurons most associated with a particular morphosyntactic category that overlap between pairs of languages. Colours in the plot refer to 2 models: \mbert (red) and \xlmrbase (blue).}
    \label{fig:violin-plot-bert-xlmr}
\end{figure}

\ryan{This paragraph need to be redone. An actual testable hypothesis is made. The idea is that the language picks up and aligns morphosyntactic markers that are \emph{functionally similar.} In a sense, this paragraph is the meat of the paper. It must be re-written with corresponding gravitas.}

In this work, we offer a concurrent hypothesis to explain the multilingual abilities of various pre-trained models; namely, that they implicitly align morphosyntactic markers that fulfil a similar grammatical function across languages, even in absence of any lexical overlap. More concretely, we conjecture that they employ the same subset of neurons to encode the same morphosyntactic information (such as gender for nouns and mood for verbs).\footnote{Concurrent work by \citet{antverg2021pitfalls} suggests a similar hypothesis based on smaller-scale experiments.}
To test the aforementioned hypothesis, we employ \citeposs{flexible-probing} latent variable probe to identify the relevant subset of neurons in each language and then measure their cross-lingual overlap.

We experiment with two multilingual pre-trained models, \mbert~\citep{devlin-etal-2019-bert} and \xlmr~\citep{conneau-etal-2020-unsupervised}, probing them for morphosyntactic information in \XX 
languages from Universal Dependencies \citep{ud-2.1}.
Based on our results, we argue that pre-trained models do indeed develop a cross-lingually entangled representation of morphosyntax.
We further note that, as the number of values of a morphosyntactic category increases,
cross-lingual alignment decreases.
Finally, we find that language pairs with high proximity (in the same genus or with similar typological features) and with vast amounts of pre-training data tend to exhibit more overlap between neurons.
Identical factors are known to affect also the empirical performance of zero-shot cross-lingual transfer \citep{wu-dredze-2019-beto}, which suggests a connection between neuron overlap and transfer abilities.

\lucas{Elaborated the end here. Previously this said ``warranting zero-shot learning for related languages'', which I don't understand. Is this what was meant?}\ryan{Expand this paragraph with actual discussion of the results.}\karolina{Expanded a bit, what do you think?}\lucas{Made some edits and merged with previous paragraph. What do you think?}

\section{Intrinsic Probing}
\label{sec:background}

Intrinsic probing aims to determine exactly which dimensions in a representation, e.g., those given by \mbert, 
encode a particular linguistic property~\citep{dalviWhatOneGrain2019,intrinsic}.
Formally, let $\Pi$ be the inventory of values that some morphosyntactic category can take in a particular language, for example $\Pi = \{\prop{fem}, \prop{msc}, \prop{neu}\}$ for grammatical gender in Russian.
Moreover, let $\calD = \{ (\pi^{(n)}, \vh^{(n)}) \}_{n=1}^N$ be a dataset of labelled embeddings such that $\pi^{(n)} \in \Pi$ and $\vh^{(n)} \in \R^d$, where $d$ is the dimensionality of the representation being considered, e.g., $d = 768$ for \mbert.
Our goal is to find a subset of $k$ neurons $C^\star \subseteq D = \{1, \ldots, d\}$, where $d$ is the total number of dimensions in the representation being probed, that maximises some informativeness measure.

In this paper, we make use of a latent-variable model recently proposed by \citet{flexible-probing} for intrinsic probing.\ryan{Give an actual description of the model.}
The idea is to train a probe with latent variable $C$ indexing the subset of the dimensions $D$ of the representation $\vh$ that should be used to predict the property $\pi$:
\begin{align}
\label{eq:joint}
    \ptheta(\pi \mid \vh) &= \sum_{C \subseteq D} \ptheta(\pi \mid \vh, C)\,p(C)
\end{align}
where we opt for a uniform prior $p(C)$ and $\vtheta$ are the parameters of the probe.

Our goal is to learn the parameters $\vtheta$. However, since the computation of \cref{eq:joint} requires us to marginalise over all subsets $C$ of $D$, which is intractable, we optimise a variational lower bound to the log-likelihood:
\begin{align}
    \NLL &(\vtheta) = \sum_{n=1}^N \log \sum_{\substack{C \subseteq D}} p_\vtheta \left(\pi^{(n)}, C \mid \vh^{(n)} \right) \, \label{eq:elbo} \\
    &\hspace{-0.2cm} \ge \sum_{n=1}^N\left( \expectq \sqr{\log \ptheta(\pi^{(n)}, C \mid \vh^{(n)})} + \entropy(\qphi)\right) \nonumber
\end{align}
where $\entropy(\cdot)$ stands for the entropy of a distribution, and $\qphi(C)$ is a variational distribution over subsets $C$.\footnote{We refer the reader to \citet{flexible-probing} for a full derivation of \cref{eq:elbo}.}
For this paper, we chose $\qphi(\cdot)$ to correspond to a Poisson sampling scheme~\citep{lohr2019sampling}, which models a subset as being sampled by 
subjecting each dimension to an independent Bernoulli trial, where $\phi_i$ parameterises the probability of sampling any given dimension.\footnote{We opt for this sampling scheme as \citet{flexible-probing} found that it is more computationally efficient than conditional Poisson~\citep{hajekAsymptoticTheoryRejective1964} while maintaining performance.}

Having trained the probe, all that remains is using it to identify the subset of dimensions that is most informative about the morphosyntactic category we are probing for.
We do so by finding the subset $C^\star_k$ of $k$ neurons maximising the posterior:
\begin{align}
    C_k^{\star} &= \argmax_{\substack{C \subseteq D, \\ |C| = k}} \log \ptheta(C \mid \calD)
    \label{eq:general-objective}
\end{align} 
In practice, this combinatorial optimisation problem is intractable. Hence, we solve it using greedy search.

\section{Experimental Setup}




We now describe the experimental methodology of the paper, including the data, training procedure and statistical testing.

\paragraph{Data.} We select \XX treebanks from Universal Dependencies 2.1~\citep[UD;][]{ud-2.1}, which contain sentences annotated with morphosyntactic information in a wide array of languages. 
Afterwards, we compute contextual representations for every individual word in the treebanks using multilingual \bert (\mbert-base)\lucas{we should say whether it's \mbert base or large} and the base and large versions of XLM-RoBERTa (\xlmrbase and \xlmrlarge). We then associate each word with its parts of speech and morphosyntactic features, which are mapped to the UniMorph schema~\citep{kirovUniMorphUniversalMorphology2018}.\footnote{We use the converter developed for UD v2.1 from~\citet{mccarthyMarryingUniversalDependencies2018}.} The selected treebanks include all languages supported by both \mbert and \xlmr which are available in UD.

Rather than adopting the default UD splits, we re-split word representations based on lemmata ending up with disjoint vocabularies for the train, development, and test set. This prevents a probe from achieving high performance by sheer memorising. 
Moreover, for every category--language pair (e.g., mood--Czech), we discard any lemma with fewer than 20 
tokens in its split.

\begin{figure}[t]
    \centering
    
    \begin{subfigure}{\columnwidth}
    \centering
    \includegraphics[width=\linewidth]{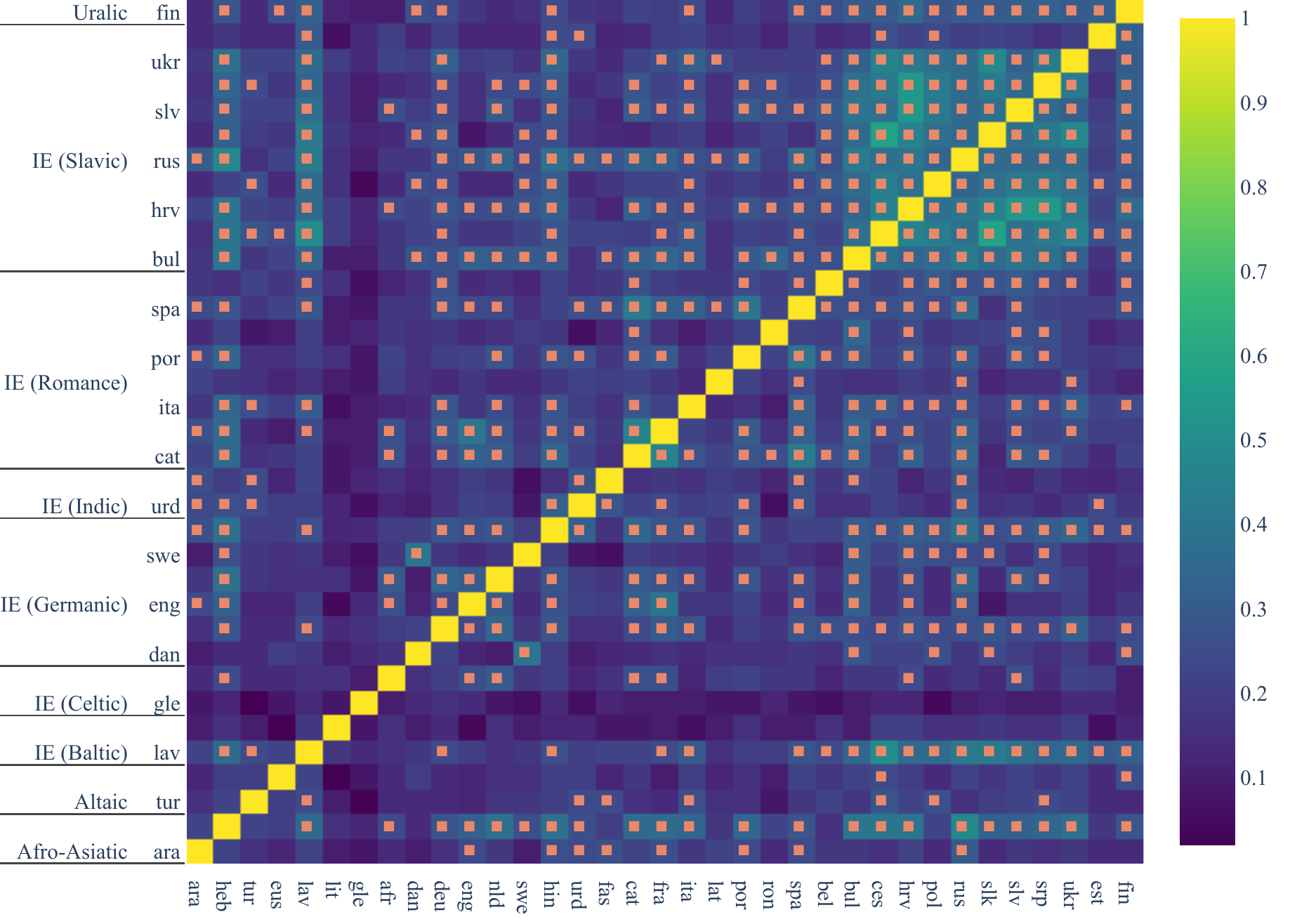}
    \end{subfigure}
    
    \begin{subfigure}{\columnwidth}
    \centering
    \includegraphics[width=\linewidth]{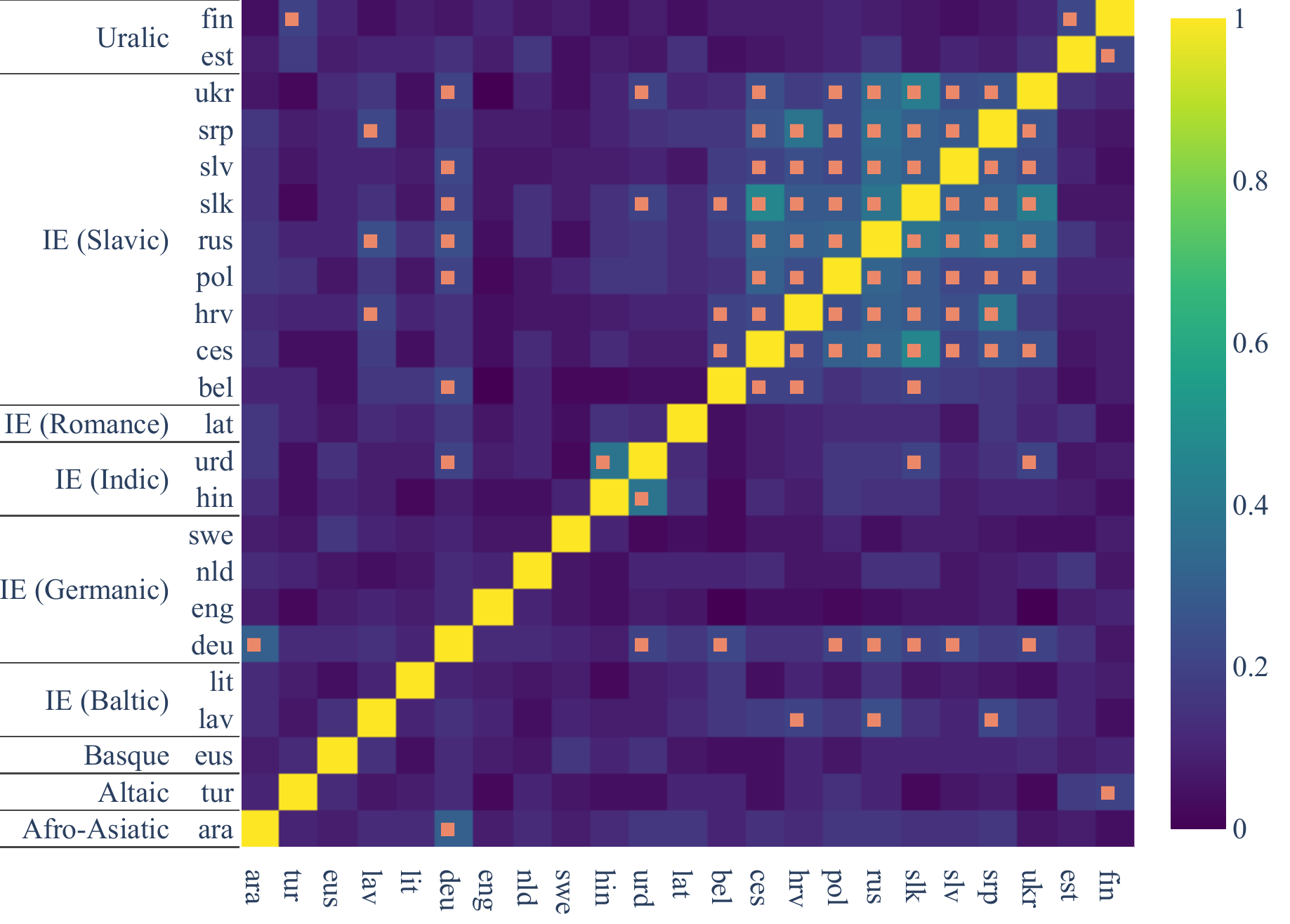}
    \end{subfigure}

    \caption{The percentage overlap between the top-50 most informative number dimensions in \mbert for number (top) and \xlmrlarge for case (bottom). 
    Statistically significant overlap after Holm--Bonferroni family-wise error correction~\citep{holmSimpleSequentiallyRejective1979}, with $\alpha = 0.05$, 
    is marked with an orange square.}
    \label{fig:neuron-overlap}
\end{figure}

\paragraph{Training.}
We first train a probe for each morphosyntactic category--language combination with the objective in \cref{eq:elbo}. In line with established practices in probing, we parameterise $\ptheta(\cdot)$ as a linear layer followed by a softmax. Afterwards, we identify the top-$k$ most informative neurons in the last layer of m-\bert, \xlmrbase, and \xlmrlarge. Specifically, following \citet{intrinsic}, we use the log-likelihood of the probe on the test set as our greedy selection criterion. We single out 50 dimensions for each combination of morphosyntactic category and language.\footnote{We select this number as a trade-off between the size of a probe and a tight estimate of the mutual information based on the results presented in \citet{flexible-probing}.}

Next, we measure the pairwise overlap in the top\nobreakdash-$k$ most informative dimensions between all pairs of languages where a morphosyntactic category is expressed. This results in matrices such as \cref{fig:neuron-overlap}, where the pair-wise percentages of overlapping dimensions are visualised as a heat map.

\paragraph{Statistical Significance.}
Suppose that two languages have $m \in \{1, \ldots, k\}$\ryan{So you type $m$ as a set of the first $k$ integers. But later in the text you treat $m$ as if it is the \emph{number} of overlapping integers. I am a confused boi.}\karolina{right, $m$ is a number from this set} overlapping neurons when considering the top-$k$ selected neurons for each of them.
To determine whether such overlap is statistically significant, we compute the probability of an overlap of \emph{at least} $m$ neurons under the null hypothesis that the sets of neurons are sampled independently at random. 
We estimate these probabilities 
with a permutation test.
In this paper, we set a threshold of $\alpha = 0.05$ for significance.

\paragraph{Family-wise Error Correction.}
Finally, we use Holm-Bonferroni~\citep{holmSimpleSequentiallyRejective1979} family-wise error correction. Hence, our threshold is appropriately adjusted for multiple comparisons, which makes incorrectly rejecting the null hypothesis less likely.

In particular, the individual permutation tests are ordered in ascending order of their $p$-values. 
The test with the smallest probability undergoes the Holm--Bonferroni correction \citep{holmSimpleSequentiallyRejective1979}.
If already the first test is not significant, the procedure stops; otherwise, the test with the second smallest $p$-value is corrected for a family of $t-1$ tests, where $t$ denotes the number of conducted tests. The procedure stops either at the first non-significant test or after iterating through all $p$-values. This sequential approach guarantees that the probability that we incorrectly reject \emph{one or more} of the hypotheses is at most $\alpha$.\looseness=-1


\section{Results}
\label{sec:results}

We first consider whether multilingual pre-trained models develop a cross-lingually entangled notion of morphosyntax: for this purpose, we measure the overlap between subsets of neurons encoding similar morphosyntactic categories across languages. 
Further, we debate whether the observed patterns are dependent on various factors, such as morphosyntactic category, language proximity, pre-trained model, and pre-training data size.   

\begin{figure}[t]
    \centering
    \includegraphics[width=\linewidth]{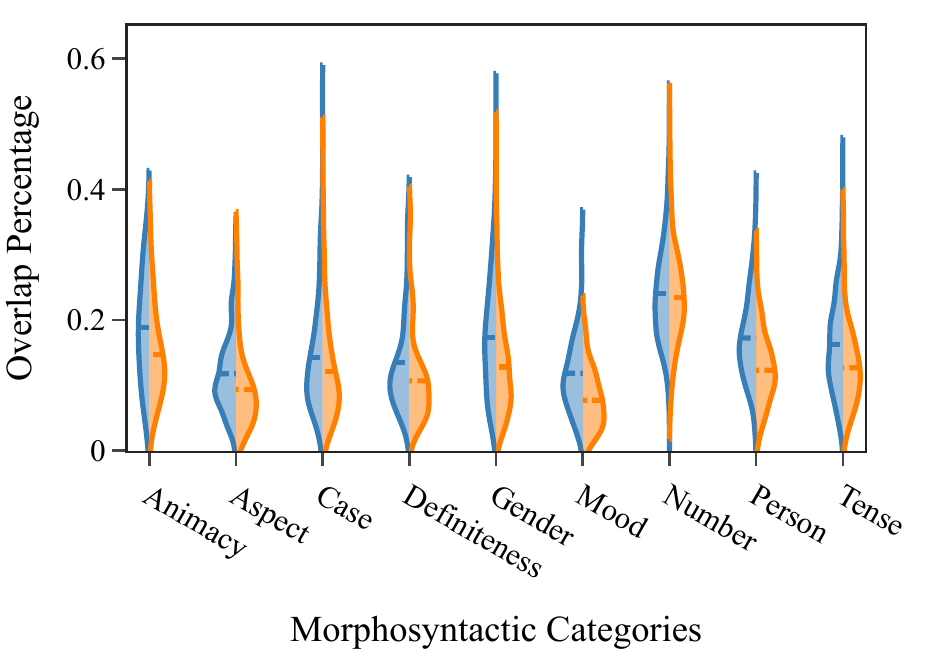}
    \caption{Ratio of neurons most associated with a particular morphosyntactic category that overlap between pairs of languages. Colours in the plot refer to 2 models: \xlmrbase (blue) and \xlmrlarge (orange).}
    \label{fig:violin-plot-xlmr-base-large}
\end{figure}

\paragraph{Neuron Overlap.}
The matrices of pairwise overlaps for each of the 14 categories, such as \cref{fig:neuron-overlap} for number and case, are reported in \cref{app:pairoverlap}. We expand upon these results in two ways. 
First, we report the cross-lingual distribution for each category in \cref{fig:violin-plot-bert-xlmr} for \mbert and \xlmrbase, and in an equivalent plot comparing \xlmrbase and \xlmrlarge in \cref{fig:violin-plot-xlmr-base-large}. Second, 
we calculate how many overlaps are statistically significant out of the total number of pairwise comparisons in \cref{tab:overlap-rates}. 
From the above results, it emerges that $\approx 20$\% of neurons among the top-$50$ most informative ones overlap on average, 
but this number may vary dramatically across categories. 

\begin{figure}[t]
    \centering
    
    \caption{Spearman's correlation, for a given model and morphological category, between the cross-lingual average percentage of overlapping neurons and:}
\label{fig:correlations-all}

\begin{subfigure}{\columnwidth}
    \centering
    \includegraphics[width=\linewidth]{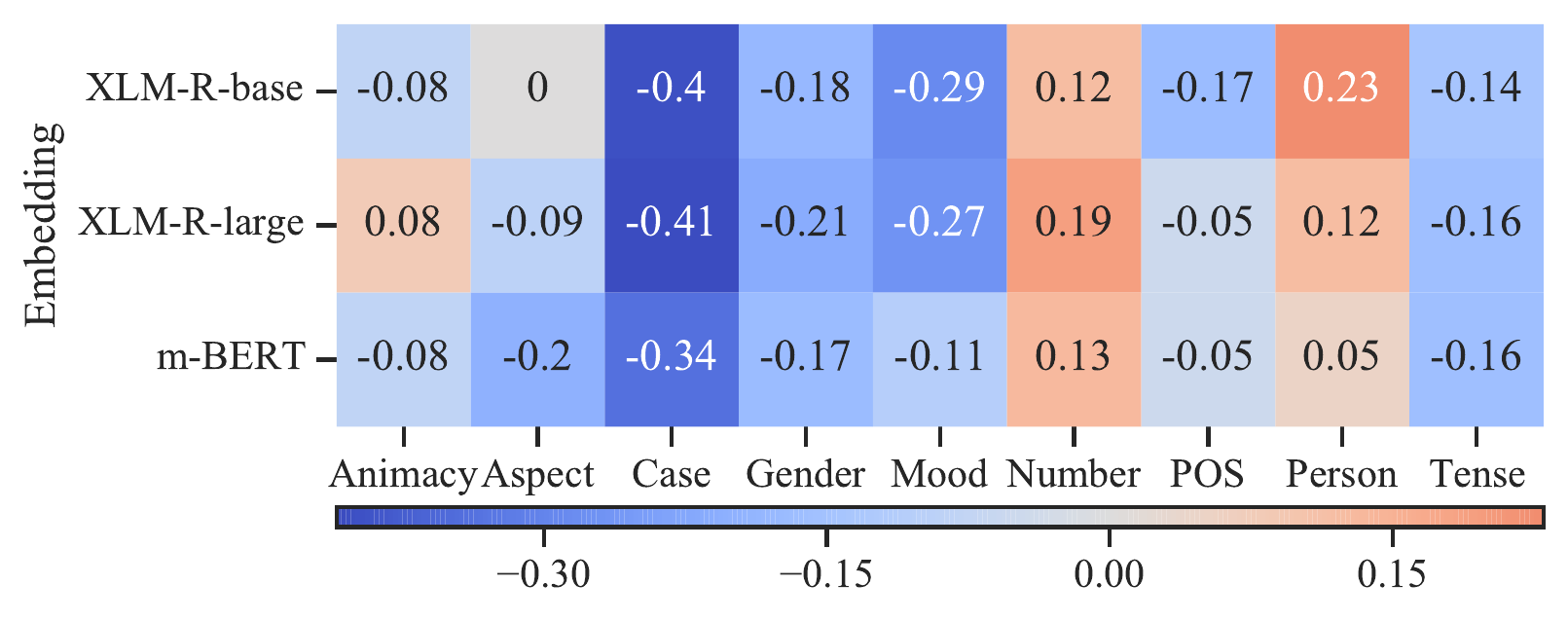}
    \caption{number of values for each morphosyntactic category;}
    \label{fig:no-attribute-overlap}
\end{subfigure}
    
\begin{subfigure}{\columnwidth}
    \centering
    \includegraphics[width=\linewidth]{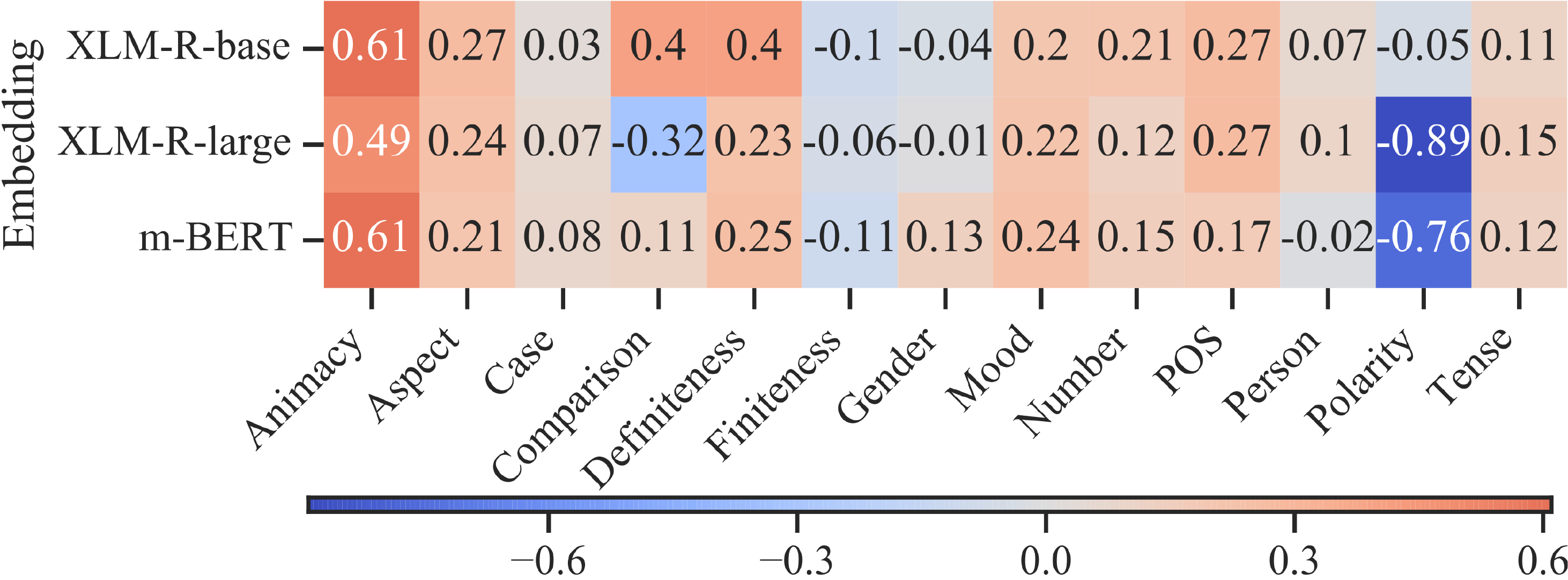}
    \caption{typological similarity;}
    \label{fig:similarity-overlap}
\end{subfigure}

\begin{subfigure}{\columnwidth}
    \centering
    \includegraphics[width=\linewidth]{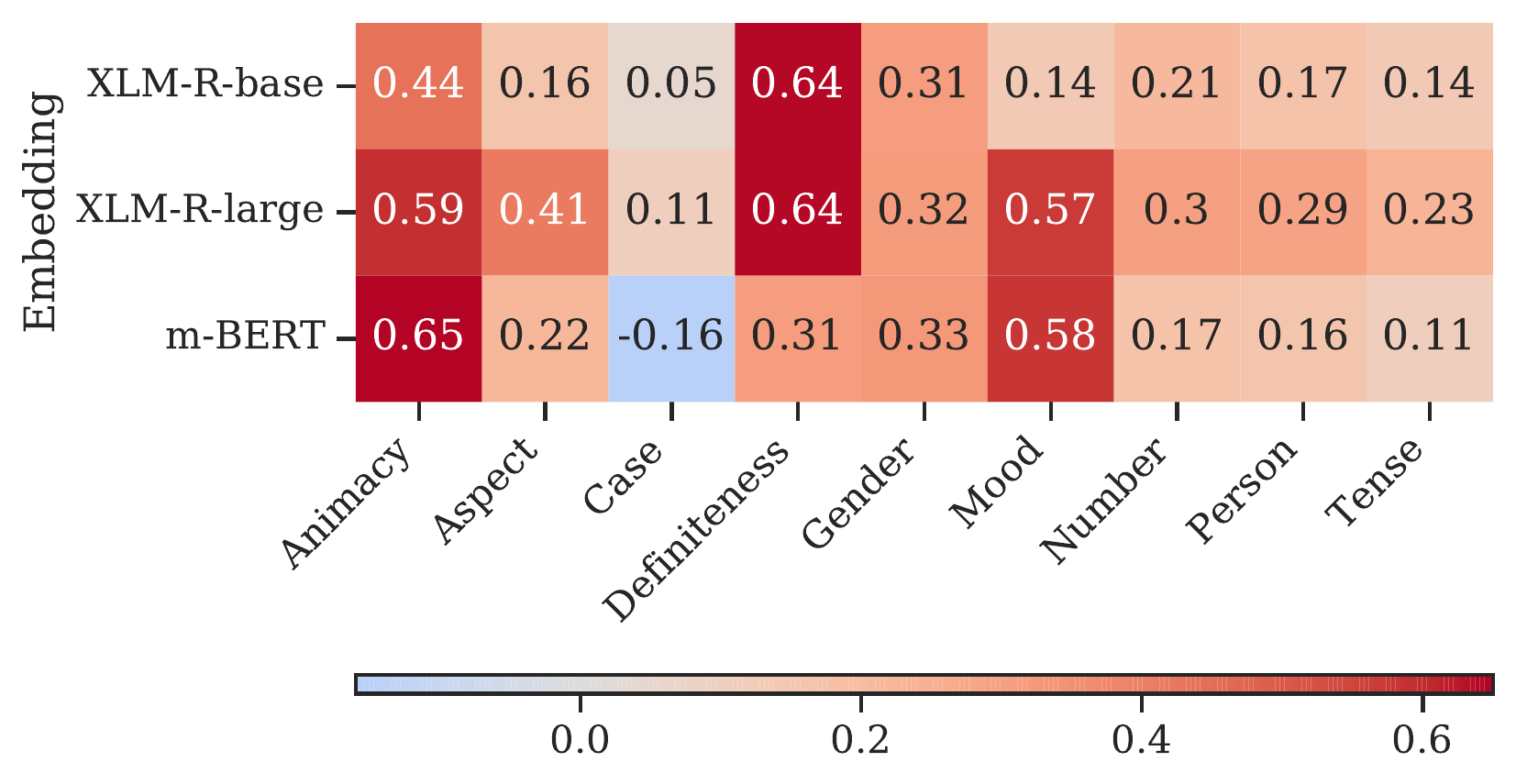}
    \caption{language model training data size.}
    \label{fig:traindatasize}
\end{subfigure}

\end{figure}

\paragraph{Morphosyntactic Categories.}
Based on \cref{tab:overlap-rates}, significant overlap is particularly accentuated in specific categories, such as comparison, polarity, and number. However, neurons for other categories such as mood, aspect, and case are shared by only a handful of language pairs despite the high number of comparisons. This finding may be partially explained by the different number of values each category can take. Hence, we test whether there is a correlation between this number and average cross-lingual overlap in \cref{fig:no-attribute-overlap}. As expected, we generally find negative correlation coefficients---prominent exceptions being number and person. As the inventory of values of a category grows, cross-lingual alignment becomes harder.

\begin{table}[t]
\centering
\begin{tabular}{lrrrr}
\toprule
{} &  \rotatebox[origin=l]{90}{m-BERT} &  \rotatebox[origin=l]{90}{XLM-R-base} &  \rotatebox[origin=l]{90}{XLM-R-large} &  \rotatebox[origin=l]{90}{Total} \\
\midrule
Definiteness &    0.11 &        0.22 &         0.13 &     45 \\
Comparison   &    0.20 &        0.90 &         0.50 &     10 \\
Possession   &    0.00 &        0.00 &         0.00 &      1 \\
Aspect       &    0.03 &        0.10 &         0.09 &    153 \\
Polarity     &    0.33 &        0.67 &         0.33 &      3 \\
Number       &    0.40 &        0.51 &         0.74 &    666 \\
Animacy      &    0.14 &        0.57 &         0.32 &     28 \\
Mood         &    0.00 &        0.07 &         0.05 &    105 \\
Gender       &    0.15 &        0.32 &         0.19 &    378 \\
Person       &    0.08 &        0.25 &         0.13 &    276 \\
POS          &    0.04 &        0.27 &         0.70 &    861 \\
Case         &    0.10 &        0.18 &         0.17 &    300 \\
Tense        &    0.08 &        0.23 &         0.12 &    325 \\
Finiteness   &    0.09 &        0.18 &         0.09 &     45 \\
\bottomrule
\end{tabular}
\caption{Proportion of language pairs with statistically significant overlap in the top-50 neurons for an attribute (after Holm--Bonferroni~\citep{holmSimpleSequentiallyRejective1979} correction). We compute these ratios for each model. The final column reports the total number of pairwise comparisons.}
\label{tab:overlap-rates}
\end{table}

\begin{figure}[t]
    \centering
    \includegraphics[width=\linewidth]{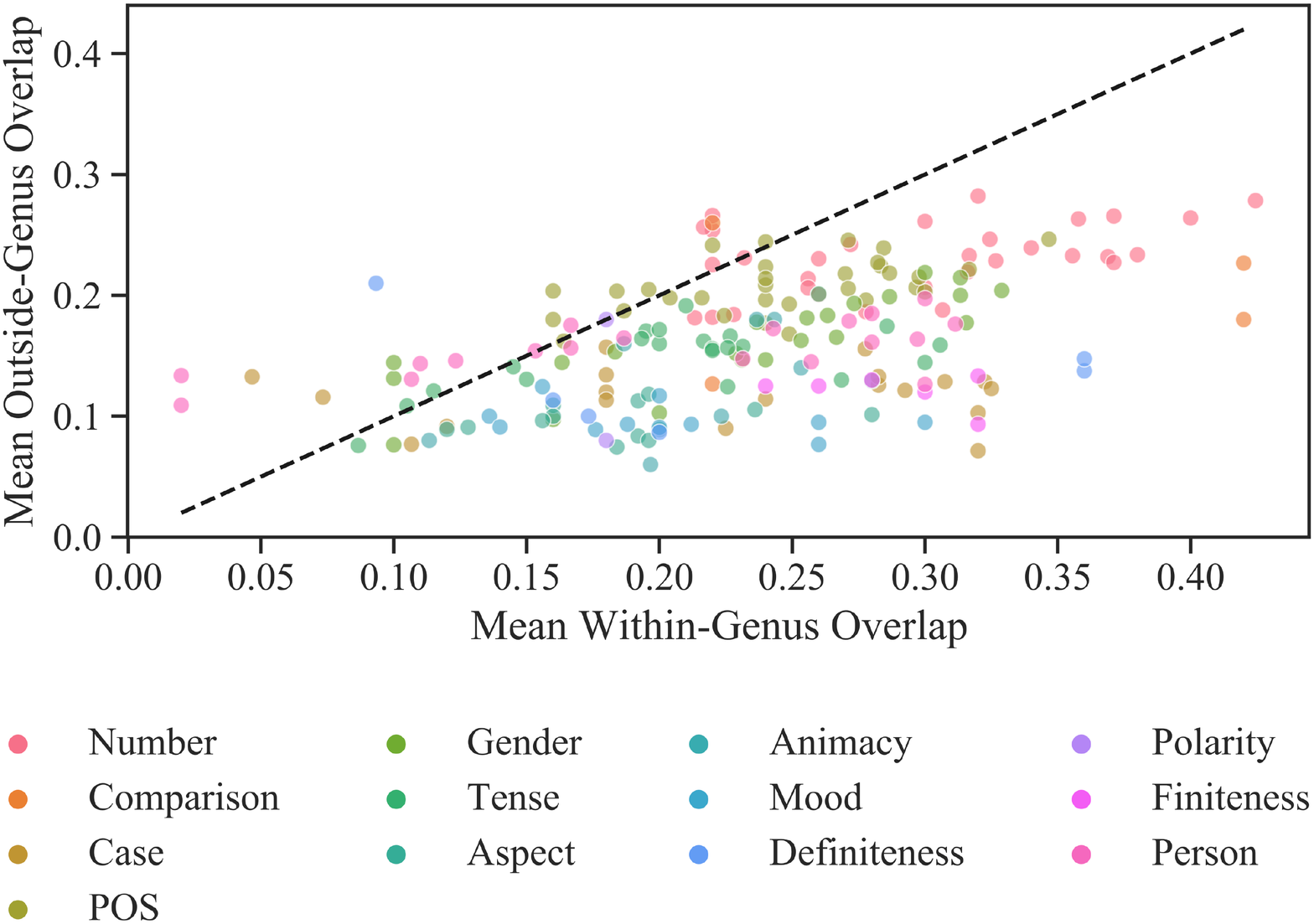}
    \caption{Mean percentage of neuron overlap in \xlmrbase with languages either within or outside the same genus for each morphosyntactic category.}
    \label{fig:xlmr-base-genus-overlap}
\end{figure}

\paragraph{Language Proximity.} 
Moreover, we investigate whether language proximity, in terms of both language family and typological features, bears any relationship with the neuron overlap for any particular pair. In \cref{fig:xlmr-base-genus-overlap}, we plot pairwise similarities with languages within the same genus (e.g., Baltic) against those outside. From the distribution of the dots, we can extrapolate that sharing of neurons is more likely to occur between languages in the same genus. This is further corroborated by the language groupings emerging in the matrices of \cref{app:pairoverlap}.



In \cref{fig:similarity-overlap}, we also measure the correlation between neuron overlap and similarity of syntactic typological features based on \citet{littell2017uriel}. While correlation coefficients are mostly positive (with the exception of polarity), we remark that the patterns are strongly influenced by whether a category is typical for a specific genus. For instance, correlation is highest for animacy, a category almost exclusive to Slavic languages in our sample. 

\paragraph{Pre-trained Models.}
Afterwards, we determine whether the 3 models under consideration reveal different patterns.
Comparing \mbert and \xlmrbase in \cref{fig:violin-plot-bert-xlmr}, we find that, on average, \xlmrbase tends to share more neurons when encoding particular morphosyntactic attributes.
Moreover, comparing \xlmrbase to \xlmrlarge in \cref{fig:violin-plot-xlmr-base-large} suggests that more neurons are shared in the former than in the latter.

Altogether, these results seem to suggest that the presence of additional training data engenders cross-lingual entanglement, but increasing model size incentivises morphosyntactic information to be allocated to different subsets of neurons.
We conjecture that this may be best viewed from the lens of compression: if model size is a bottleneck, then, to attain good performance across many languages, a model is forced to learn cross-lingual abstractions that can be reused.

\paragraph{Pre-training Data Size.} Finally, we assess the effect of pre-training data size\footnote{We rely on the CC-100 statistics reported by \citet{conneau-etal-2020-unsupervised} for \xlmr and on the Wikipedia dataset's size with TensorFlow datasets \citep{tensorflow} for \mbert.} for neuron overlap. According to \cref{fig:traindatasize}, their correlation is very high. We explain this phenomenon with the fact that more data yields higher-quality (and as a consequence, more entangled) multilingual representations.

\section{Conclusions}

In this paper, we hypothesise that the ability of multilingual models to generalise across languages results from cross-lingually entangled representations, where the same subsets of neurons encode universal morphosyntactic information. We validate this claim with a large-scale empirical study on \XX 
languages and 3 models, \mbert, \xlmrbase, and \xlmrlarge. 
We conclude that the overlap is statistically significant for a notable amount of language pairs for the considered attributes. However, the extent of the overlap varies across morphosyntactic categories and tends to be lower for categories with large inventories of possible values. Moreover, we find that neuron subsets are shared mostly between languages in the same genus or with similar typological features. Finally, we discover that the overlap of each language grows proportionally to its pre-training data size, but it also decreases in larger model architectures.

Given that this implicit morphosyntactic alignment may affect the transfer capabilities of pre-trained models, we speculate that, in future work, artificially encouraging a tighter neuron overlap might facilitate zero-shot cross-lingual inference to low-resource and typologically distant languages\citep{zhao-etal-2021-inducing}.


\section*{Ethics Statement}
The authors foresee no ethical concerns with the work presented in this paper.

\section*{Acknowledgments}
This work is mostly funded by Independent Research Fund Denmark under grant agreement number 9130-00092B, as well as by a project grant from the Swedish Research Council under grant agreement number 2019-04129.
Lucas Torroba Hennigen acknowledges funding from the Michael Athans Fellowship fund.
Ryan Cotterell acknowledges support from the Swiss National Science Foundation (SNSF) as part of the ``The Forgotten Role of Inductive Bias in Interpretability'' project.

\lucas{Check bibliography}

\bibliography{lucas-references,misc_naacl}
\bibliographystyle{acl_natbib}

\clearpage
\appendix


\section{Probed Property--Language Pairs}
\label{app:probed-pairs}

\paragraph{Afro-Asiatic}
\begin{itemize}[noitemsep,topsep=0pt,parsep=0pt,partopsep=0pt]
\item\textbf{ara (Arabic)}: Gender, Voice, Mood, Part of Speech, Aspect, Person, Number, Case, Definiteness
\item\textbf{heb (Hebrew)}: Part of Speech, Number, Tense, Person, Voice    
\end{itemize}

\vspace*{-0.5\baselineskip}
\paragraph{Austroasiatic}
\begin{itemize}[noitemsep,topsep=0pt,parsep=0pt,partopsep=0pt]
 \item\textbf{vie (Vietnamese)}: Part of Speech       
\end{itemize}

\vspace*{-0.5\baselineskip}
\paragraph{Dravidian}
\begin{itemize}[noitemsep,topsep=0pt,parsep=0pt,partopsep=0pt]
  \item\textbf{tam (Tamil)}: Part of Speech, Number, Gender, Case, Person, Finiteness, Tense  
\end{itemize}

\vspace*{-0.5\baselineskip}
\paragraph{Indo-European}
\begin{itemize}[noitemsep,topsep=0pt,parsep=0pt,partopsep=0pt]
 \item \textbf{afr (Afrikaans)}: Part of Speech, Number, Tense                                              \item\textbf{bel (Berlarusian)}: Part of Speech, Tense, Number, Aspect, Finiteness, Voice, Gender, Animacy, Case, Person                            
 \item\textbf{bul (Bulgarian)}: Part of Speech, Definiteness, Gender, Number, Mood, Tense, Person, Voice, Comparison                               
 \item\textbf{cat (Catalan)}: Gender, Number, Part of Speech, Tense, Mood, Person, Aspect                                                        
 \item\textbf{ces (Czech)}: Part of Speech, Number, Case, Comparison, Gender, Mood, Person, Tense, Aspect, Polarity, Animacy, Possession, Voice
 \item\textbf{dan (Danish)}: Part of Speech, Number, Gender, Definiteness, Voice, Tense, Mood, Comparison                                       
 \item\textbf{deu (German)}: Part of Speech, Case, Number, Tense, Person, Comparison                                                                           
 \item\textbf{ell (Greek)}: Part of Speech, Case, Gender, Number, Finiteness, Person, Tense, Aspect, Mood, Voice, Comparison                   
 \item\textbf{eng (English)}: Part of Speech, Number, Tense, Case, Comparison                                                                                   
 \item\textbf{fas (Persian)}: Number, Part of Speech, Tense, Person, Mood, Comparison                                                                           
 \item\textbf{fra (French)}: Part of Speech, Number, Gender, Tense, Mood, Person, Polarity, Aspect                                              
 \item\textbf{gle (Irish)}: Tense, Mood, Part of Speech, Number, Person, Gender, Case                                                          
 \item\textbf{glg (Galician)}: Part of Speech                                                                                                                    
 \item\textbf{hin (Hindi)}: Person, Case, Part of Speech, Number, Gender, Voice, Aspect, Mood, Finiteness, Politeness                          
 \item\textbf{hrv (Croatian)}: Case, Gender, Number, Part of Speech, Person, Finiteness, Mood, Tense, Animacy, Definiteness, Comparison, Voice    
 \item\textbf{ita (Italian)}: Part of Speech, Number, Gender, Person, Mood, Tense, Aspect                                                       
 \item\textbf{lat (Latin)}: Part of Speech, Number, Gender, Case, Tense, Person, Mood, Aspect, Comparison                                      
 \item\textbf{lav (Latvian)}: Part of Speech, Case, Number, Tense, Mood, Person, Gender, Definiteness, Aspect, Comparison, Voice                 
 \item\textbf{lit (Lithuanian)}: Tense, Voice, Number, Part of Speech, Finiteness, Mood, Polarity, Person, Gender, Case, Definiteness               
 \item\textbf{mar (Marathi)}: Case, Gender, Number, Part of Speech, Person, Aspect, Tense, Finiteness                                            
 \item\textbf{nld (Dutch)}: Person, Part of Speech, Number, Gender, Finiteness, Tense, Case, Comparison                                        
 \item\textbf{pol (Polish)}: Part of Speech, Case, Number, Animacy, Gender, Aspect, Tense, Person, Polarity, Voice                              
 \item\textbf{por (Portuguese)}: Part of Speech, Person, Mood, Number, Tense, Gender, Aspect                                                        
 \item\textbf{ron (Romanian)}: Definiteness, Number, Part of Speech, Person, Aspect, Mood, Case, Gender, Tense                                    
 \item\textbf{rus (Russian)}: Part of Speech, Case, Gender, Number, Animacy, Tense, Finiteness, Aspect, Person, Voice, Comparison                
 \item\textbf{slk (Slovak)}: Part of Speech, Gender, Case, Number, Aspect, Polarity, Tense, Voice, Animacy, Finiteness, Person, Mood, Comparison
 \item\textbf{slv (Slovenian)}: Number, Gender, Part of Speech, Case, Mood, Person, Finiteness, Aspect, Animacy, Definiteness, Comparison          
 \item\textbf{spa (Spanish)}: Part of Speech, Tense, Aspect, Mood, Number, Person, Gender                                                        
 \item\textbf{srp (Serbian)}: Number, Part of Speech, Gender, Case, Person, Tense, Definiteness, Animacy, Comparison                             
 \item\textbf{swe (Swedish)}: Part of Speech, Gender, Number, Definiteness, Case, Tense, Mood, Voice, Comparison                                 
 \item\textbf{ukr (Ukrainian)}: Case, Number, Part of Speech, Gender, Tense, Animacy, Person, Aspect, Voice, Comparison                            
 \item\textbf{urd (Urdu)}: Case, Number, Part of Speech, Person, Finiteness, Voice, Mood, Politeness, Aspect

\end{itemize}

\vspace*{-0.5\baselineskip}
\paragraph{Japonic}
\begin{itemize}[noitemsep,topsep=0pt,parsep=0pt,partopsep=0pt]
      \item\textbf{jpn (Japanese)}: Part of Speech
\end{itemize}

\vspace*{-0.5\baselineskip}
\paragraph{Language isolate}
\begin{itemize}[noitemsep,topsep=0pt,parsep=0pt,partopsep=0pt]
     \item\textbf{eus (Basque)}: Part of Speech, Case, Animacy, Definiteness, Number, Argument Marking, Aspect, Comparison  
\end{itemize}

\vspace*{-0.5\baselineskip}
\paragraph{Sino-Tibetan}
\begin{itemize}[noitemsep,topsep=0pt,parsep=0pt,partopsep=0pt]
     \item\textbf{zho (Chinese)}: Part of Speech    
\end{itemize}

\vspace*{-0.5\baselineskip}
\paragraph{Turkic}
\begin{itemize}[noitemsep,topsep=0pt,parsep=0pt,partopsep=0pt]
 \item\textbf{tur (Turkish)}: Case, Number, Part of Speech, Aspect, Person, Mood, Tense, Polarity, Possession, Politeness      
\end{itemize}

\vspace*{-0.5\baselineskip}
\paragraph{Uralic}
\begin{itemize}[noitemsep,topsep=0pt,parsep=0pt,partopsep=0pt]
    \item\textbf{est (Estonian)}: Part of Speech, Mood, Finiteness, Tense, Voice, Number, Person, Case    
      \item\textbf{fin (Finnish)}: Part of Speech, Case, Number, Mood, Person, Voice, Tense, Possession, Comparison       
\end{itemize}


\section{Pairwise Overlap by Morphosyntactic Category}
\label{app:pairoverlap}

\begin{figure}[!h]
    \centering
    \caption{The percentage overlap between the top-50 most informative dimensions in a randomly selected language model for each of the morphosyntactic categories. Statistically significant overlap 
    is marked with an orange square.}
    \label{fig:attribute-overlap}
    \begin{subfigure}{\columnwidth}
    \includegraphics[width=\linewidth]{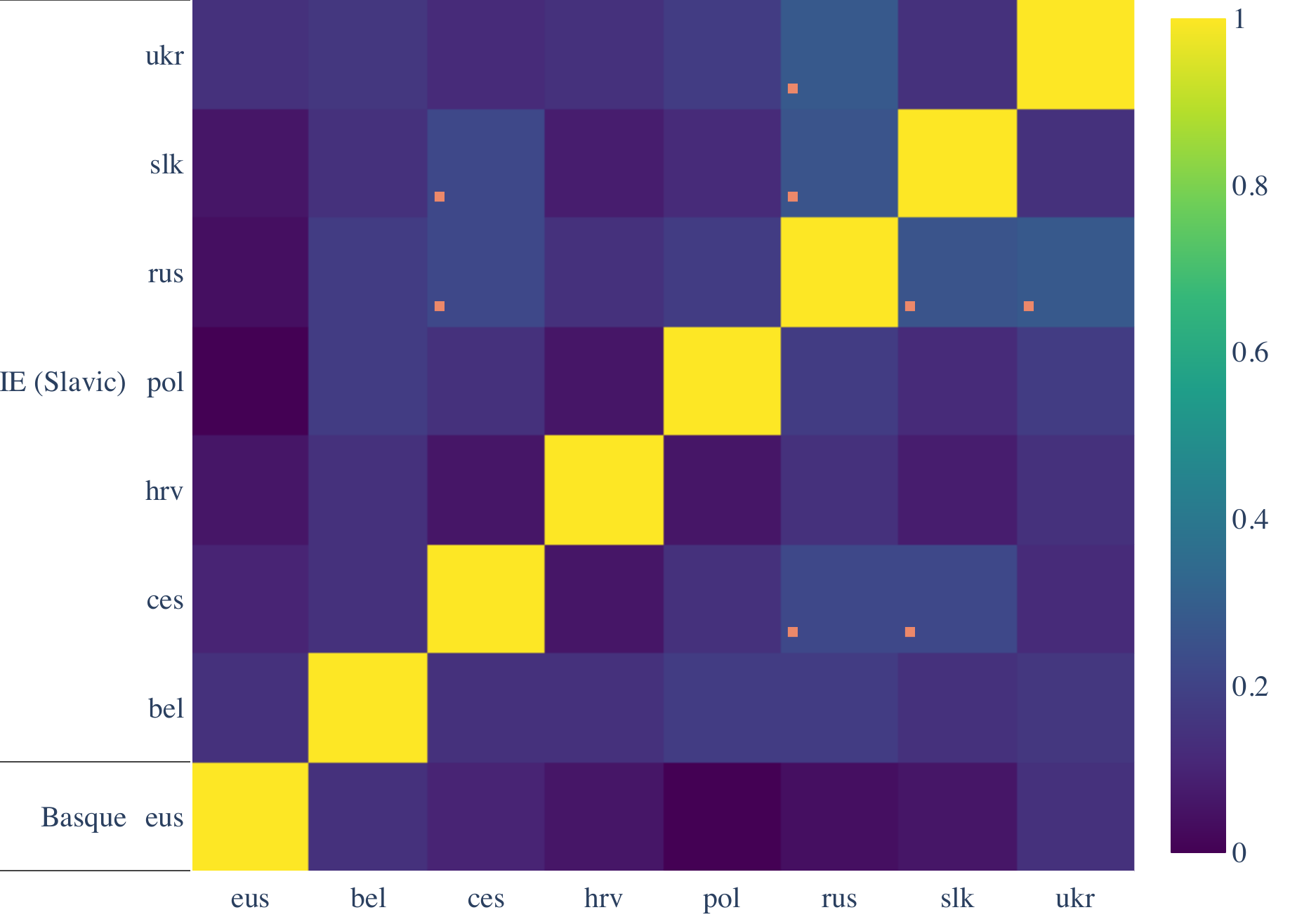}
    \caption{Animacy--\mbert}
    \label{fig:overlap-animacy}
    \end{subfigure}

\begin{subfigure}{\columnwidth}
    \centering
    \includegraphics[width=\linewidth]{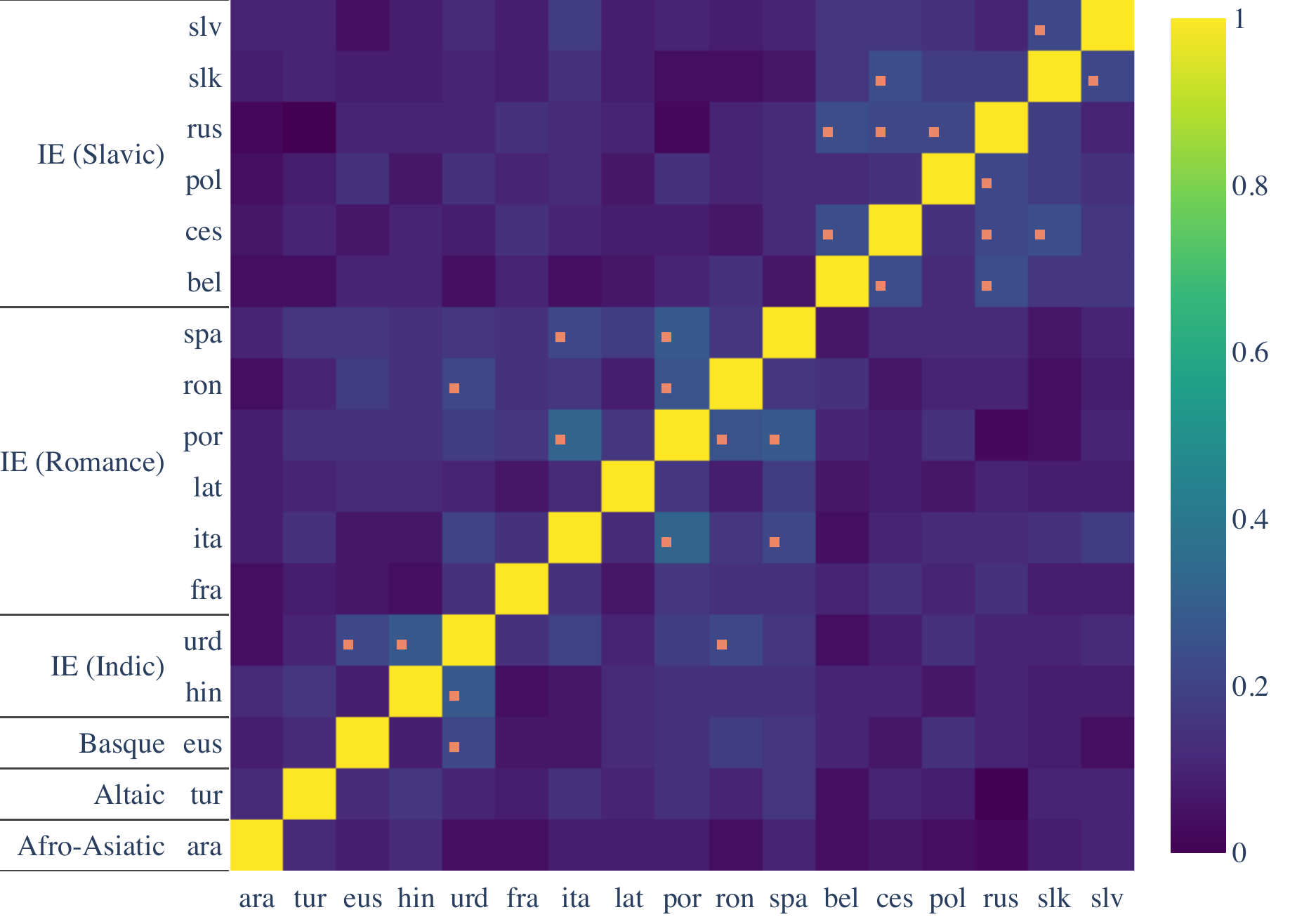}
    \caption{Aspect--\xlmrbase}
    \label{fig:overlap-aspect}
\end{subfigure}

    \vspace{-10mm}
\end{figure}

\begin{figure}[t]\ContinuedFloat
    \centering

\begin{subfigure}{\columnwidth}
    \centering
    \includegraphics[width=\linewidth]{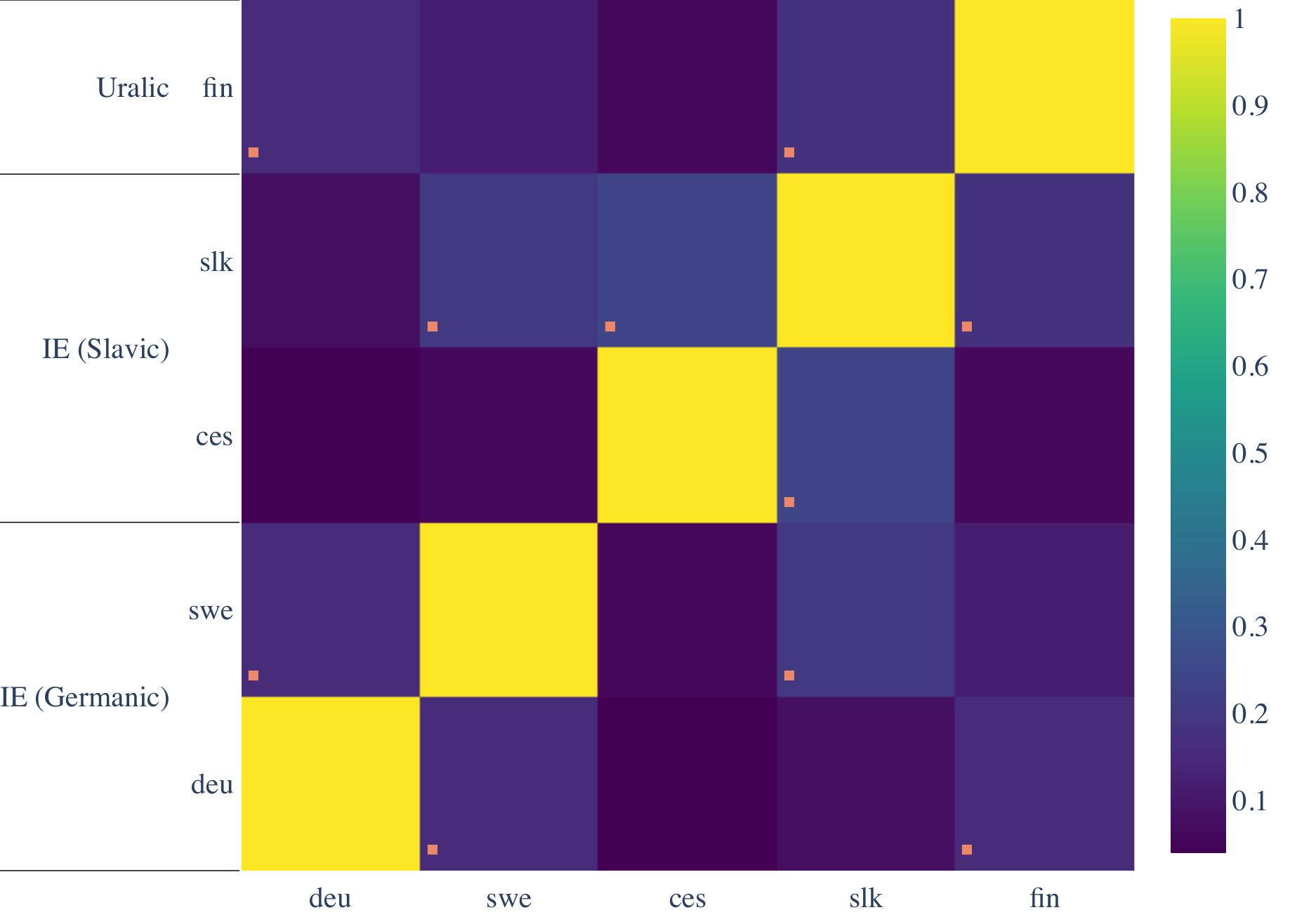}
    \caption{Comparison--\xlmrlarge}
    \label{fig:overlap-comparison}
\end{subfigure}

\begin{subfigure}{\columnwidth}
    \centering
    \includegraphics[width=\linewidth]{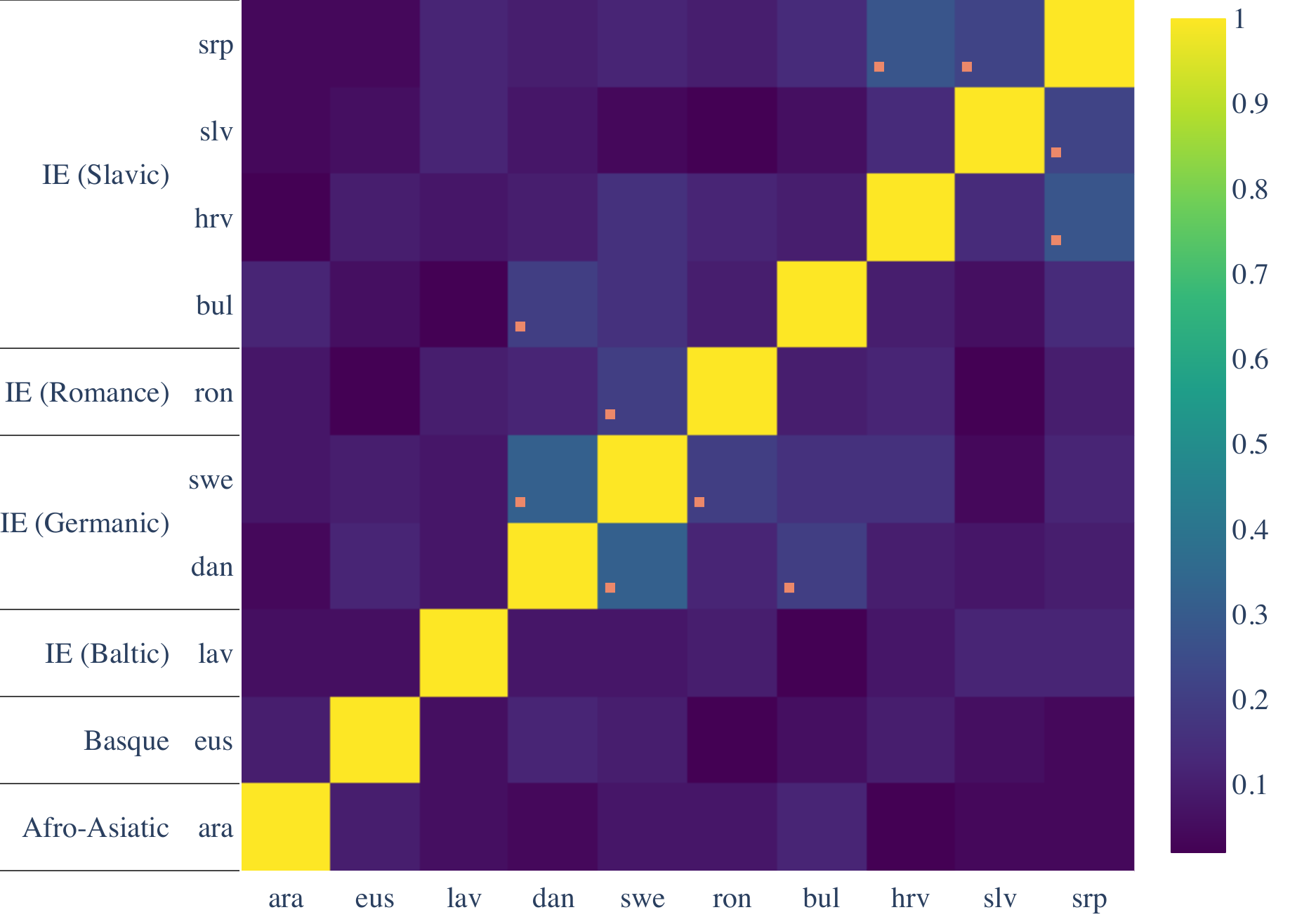}
   \caption{Definiteness--\mbert}
    \label{fig:overlap-definiteness}
\end{subfigure}

\begin{subfigure}{\columnwidth}
    \centering
    \includegraphics[width=\linewidth]{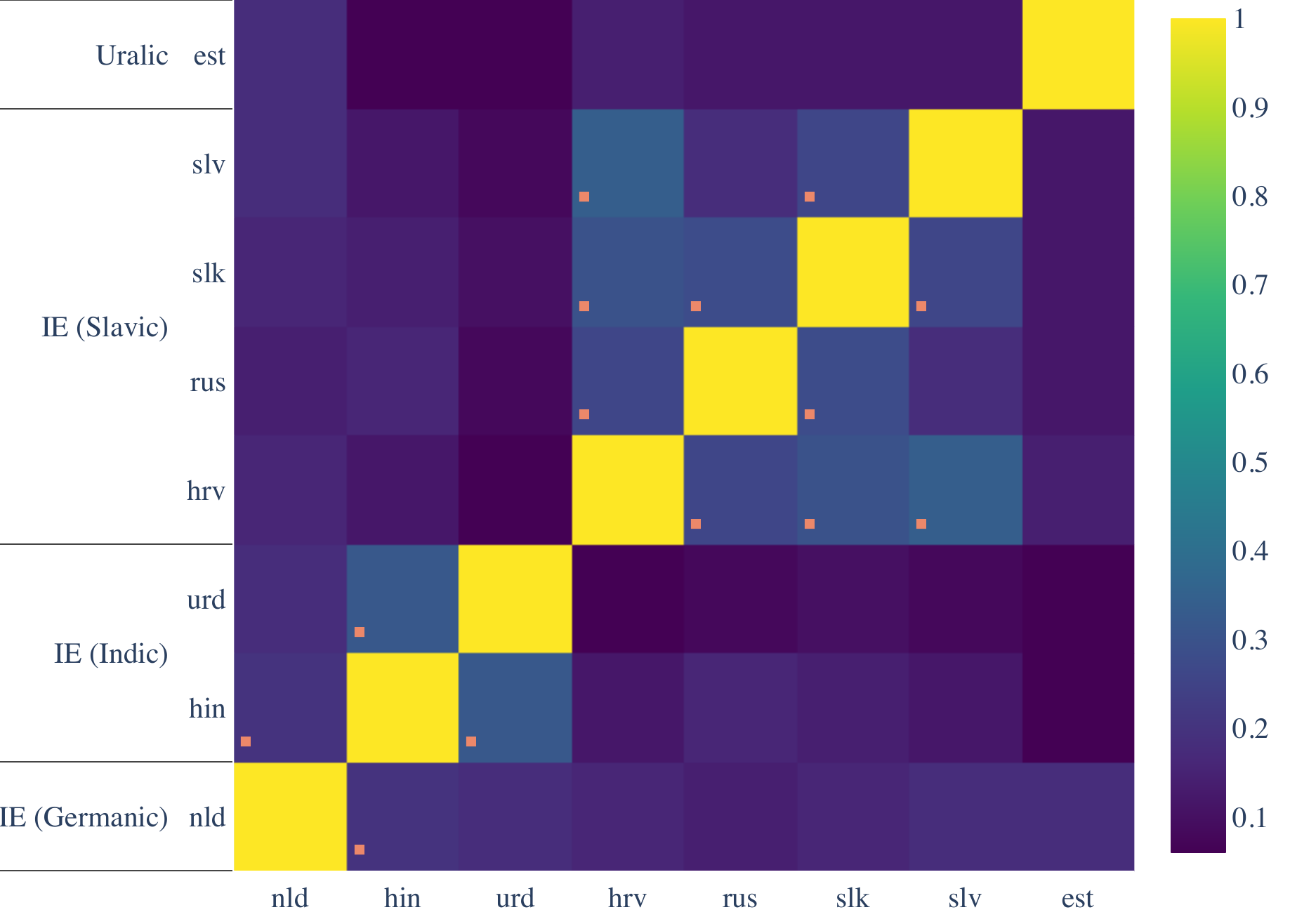}
   \caption{Finiteness--\xlmrbase}
    \label{fig:overlap-finiteness}
\end{subfigure}

\begin{subfigure}{\columnwidth}
    \centering
    \includegraphics[width=\linewidth]{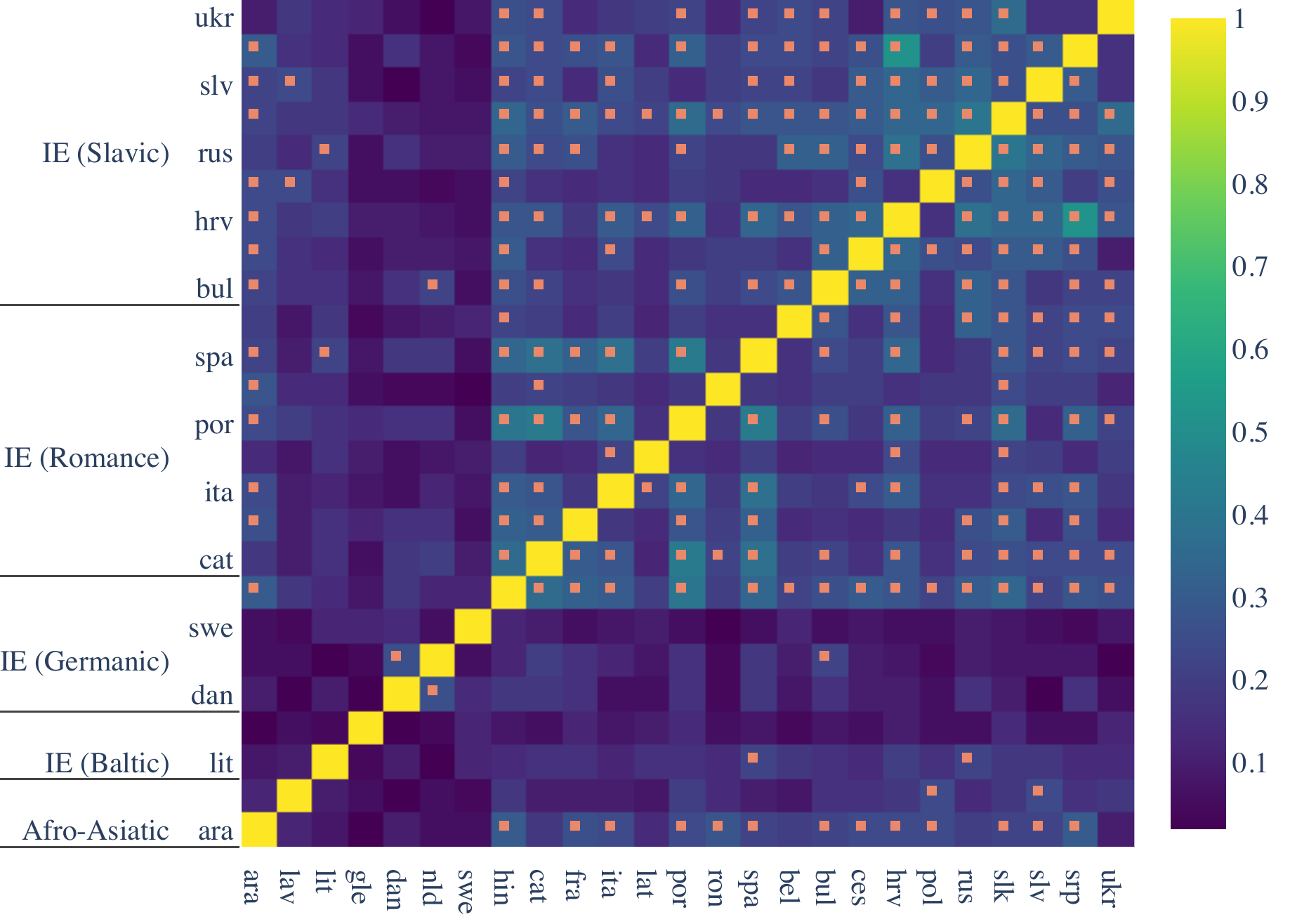}
   \caption{Gender--\xlmrbase}
    \label{fig:overlap-gender}
\end{subfigure}

    \vspace{-10mm}
\end{figure}

\begin{figure}[t]\ContinuedFloat
    \centering

\begin{subfigure}{\columnwidth}
    \centering
    \includegraphics[width=\linewidth]{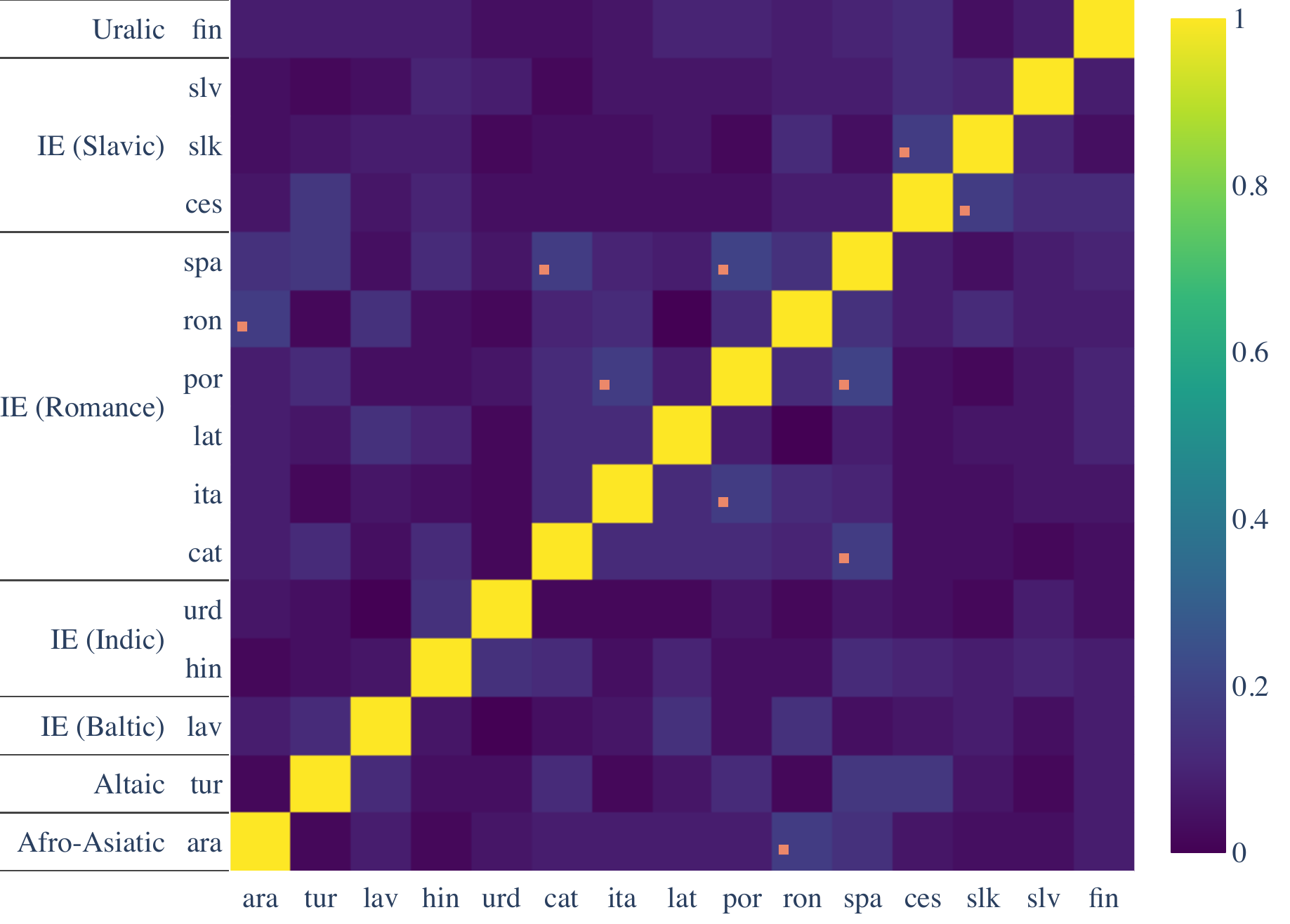}
    \caption{Mood--\xlmrlarge}
    \label{fig:overlap-mood}
\end{subfigure}

\begin{subfigure}{\columnwidth}
    \centering
    \includegraphics[width=\linewidth]{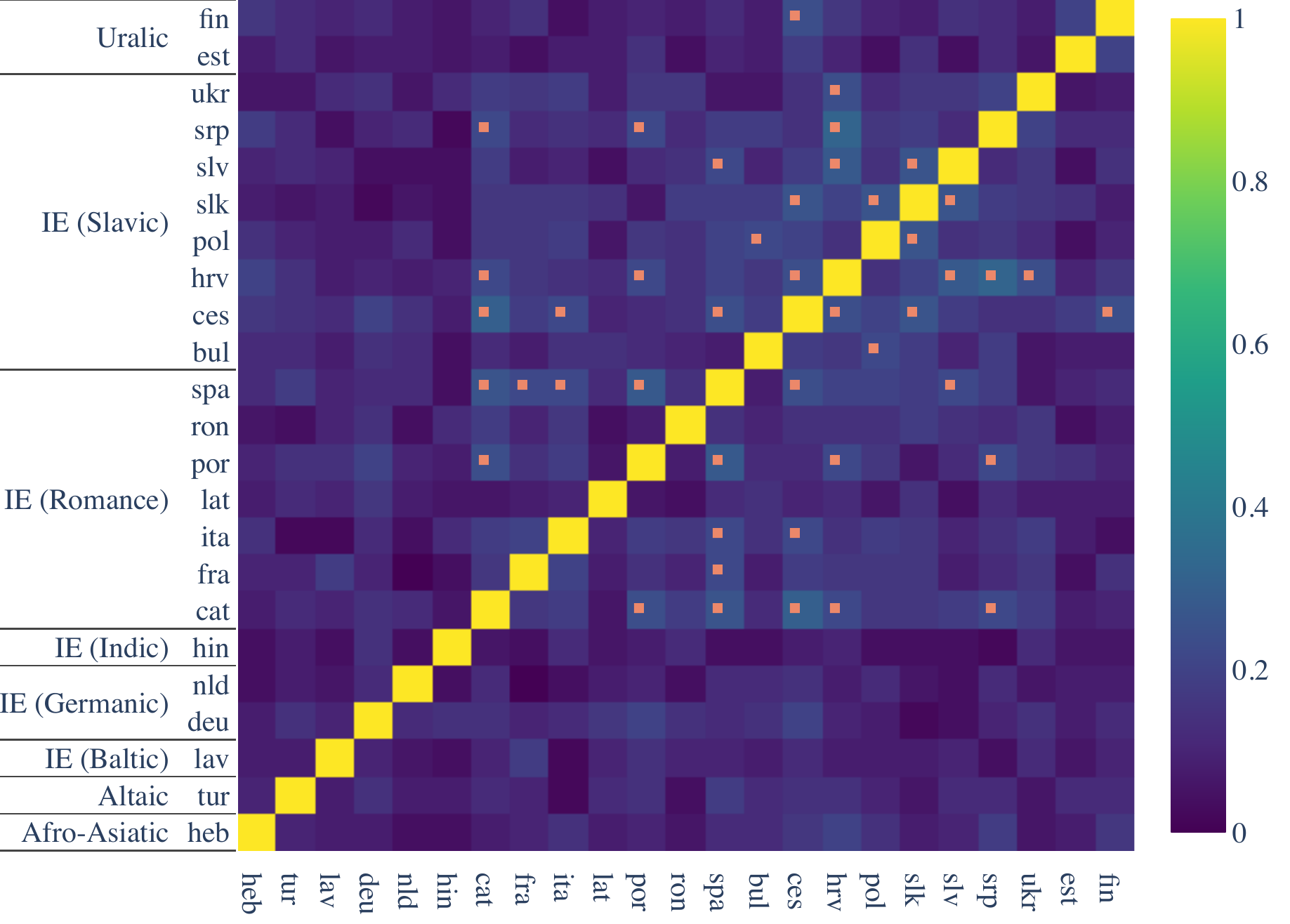}
   \caption{Person--\mbert}
    \label{fig:overlap-person}
\end{subfigure}

\begin{subfigure}{\columnwidth}
    \centering
    \includegraphics[width=\linewidth]{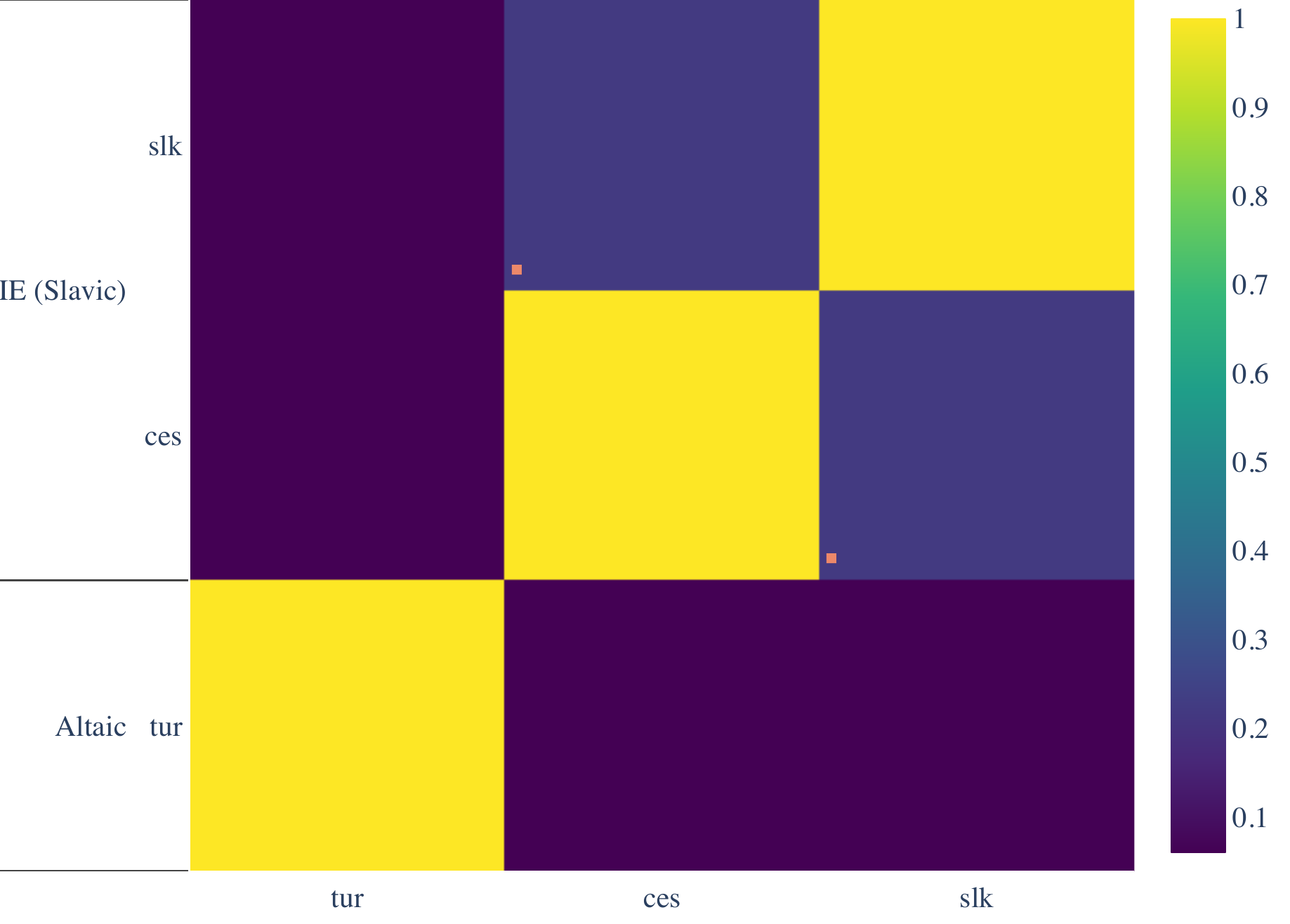}
   \caption{Polarity--\xlmrlarge}
    \label{fig:overlap-polarity}
\end{subfigure}

    \vspace{-10mm}
\end{figure}

\begin{figure}[t]\ContinuedFloat
    \centering
    
\begin{subfigure}{\columnwidth}
    \centering
    \includegraphics[width=\linewidth]{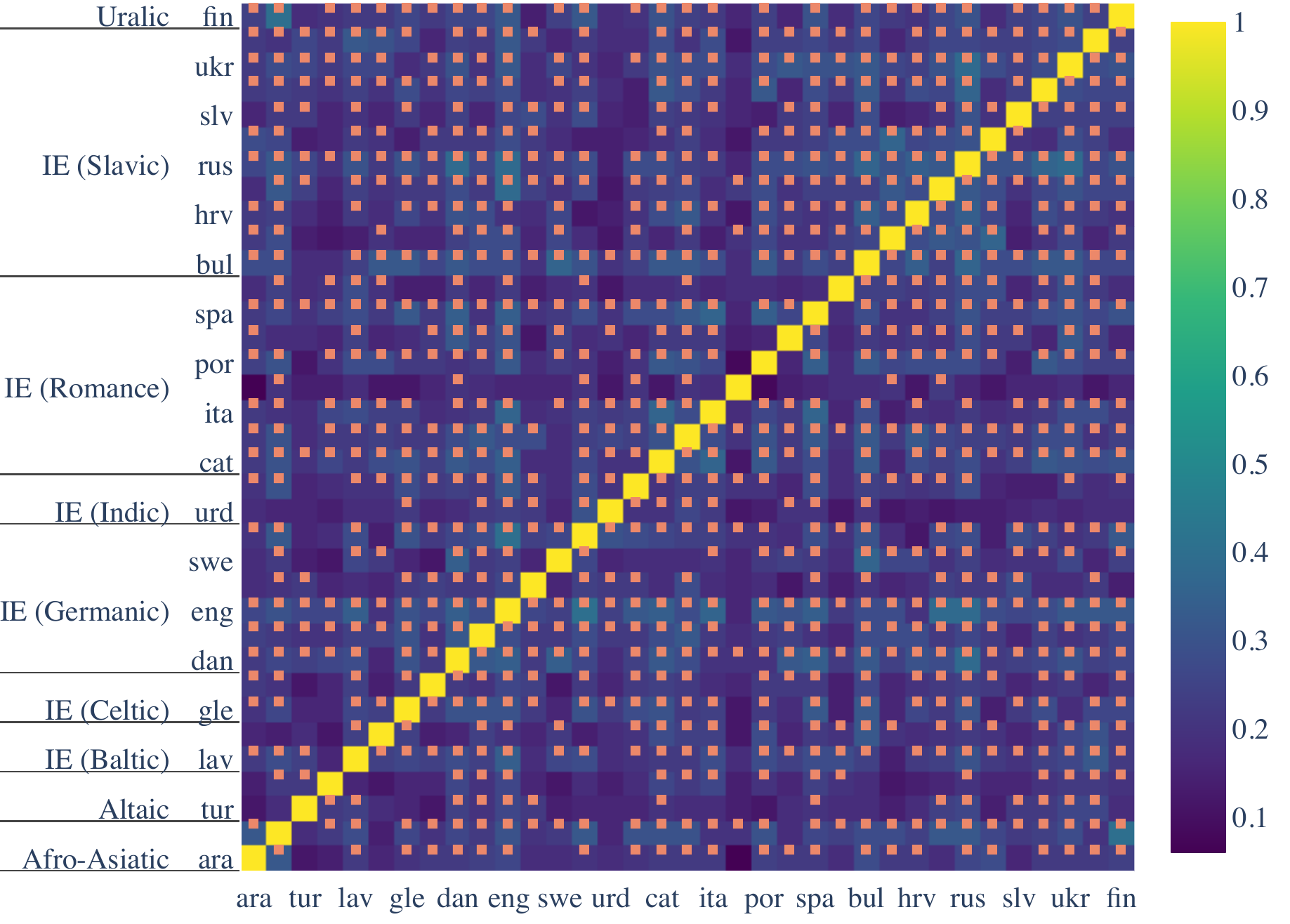}
   \caption{Part of Speech--\xlmrlarge}
    \label{fig:overlap-pos}
\end{subfigure}

\begin{subfigure}{\columnwidth}
    \centering
    \includegraphics[width=\linewidth]{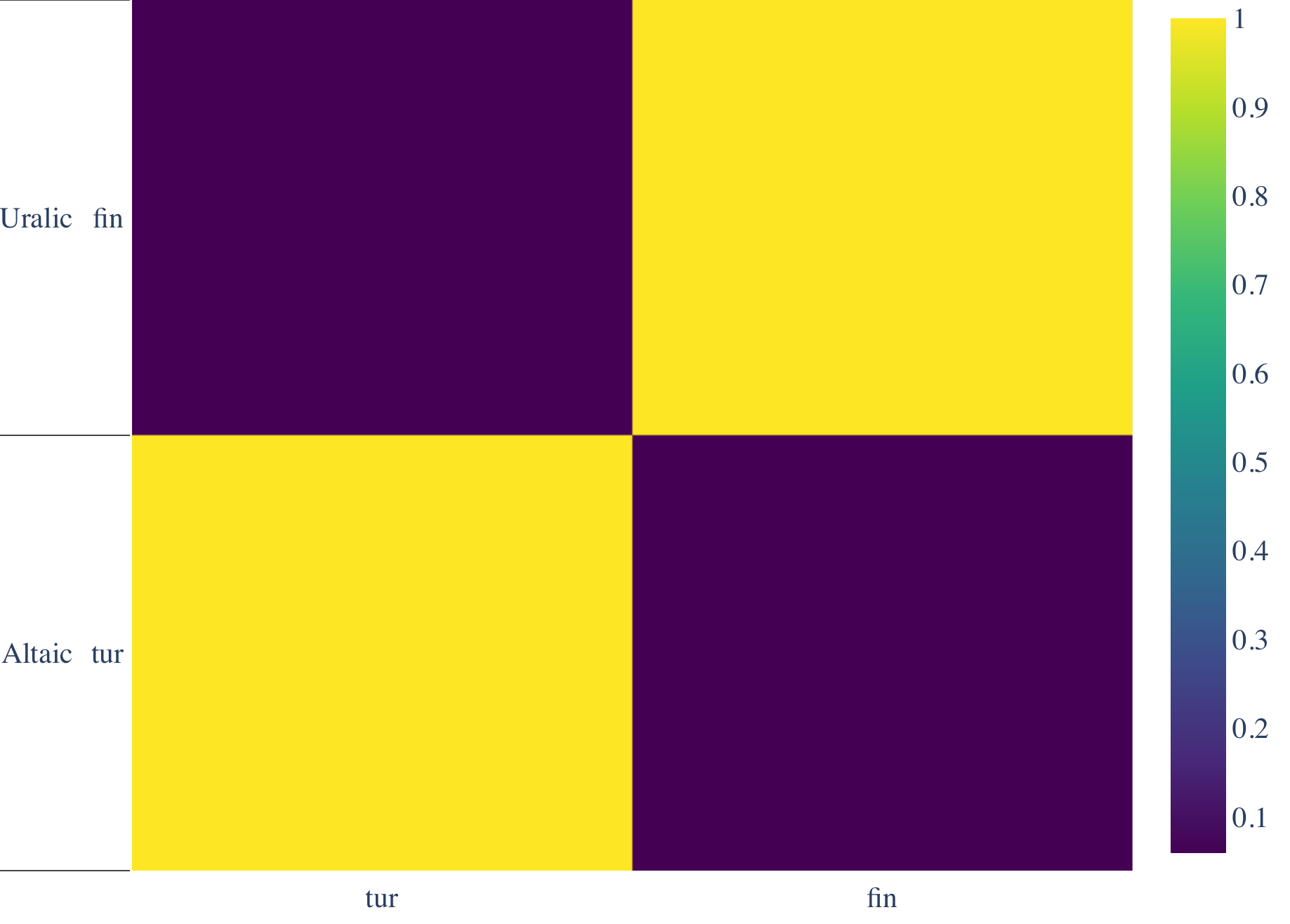}
    \caption{Possession--\xlmrbase}
    \label{fig:overlap-possession}
\end{subfigure}

\begin{subfigure}{\columnwidth}
    \centering
    \includegraphics[width=\linewidth]{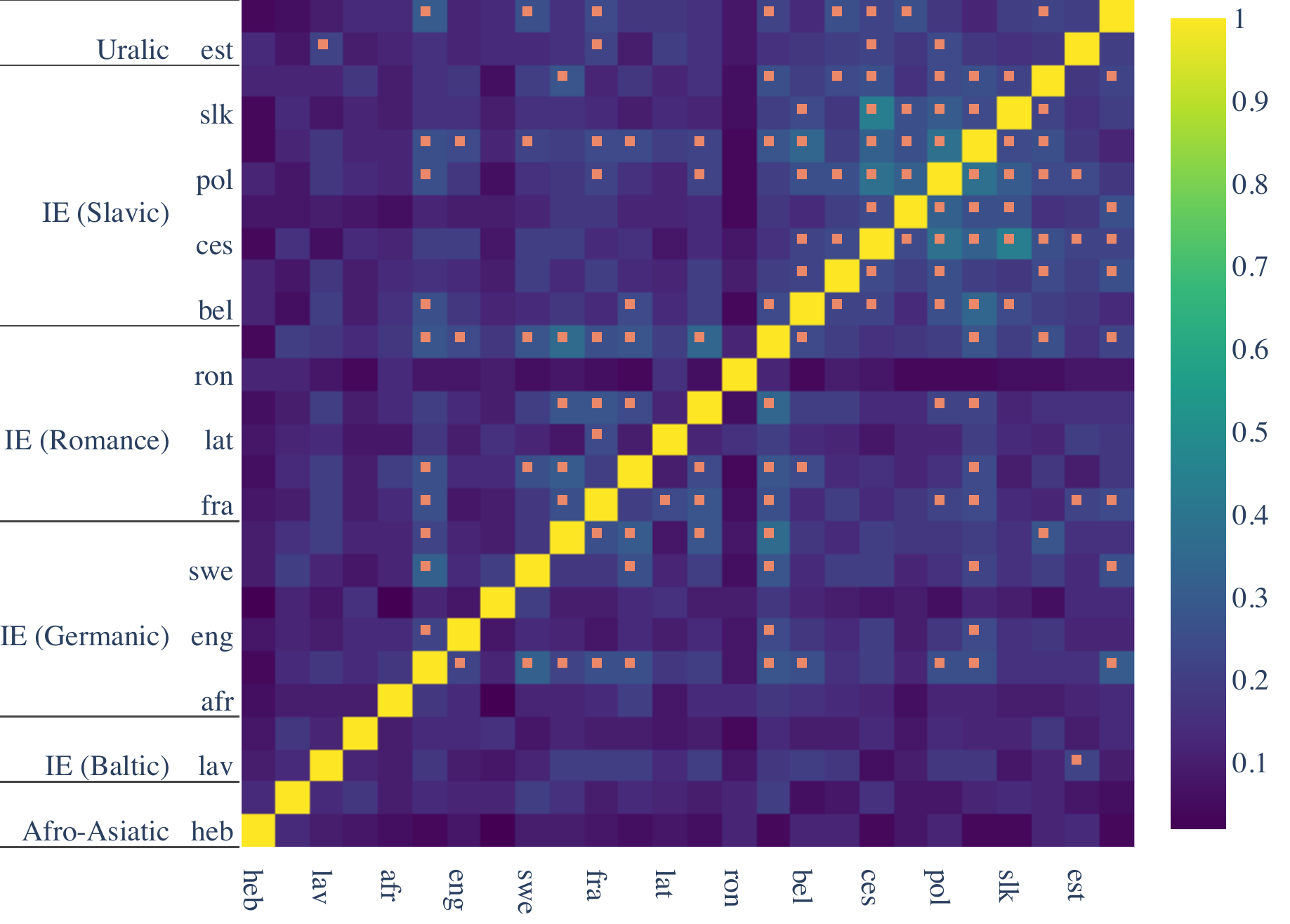}
   \caption{Tense--\xlmrbase}
    \label{fig:overlap-tense}
\end{subfigure}

    \vspace{-10mm}
\end{figure}

\end{document}

d pe

%% file: abbrevs.tex
\usepackage[T5,T1]{fontenc}
\usepackage{combelow}
\usepackage[utf8]{inputenc}

\DeclareTextSymbolDefault{\ohorn}{T5}
\DeclareTextSymbolDefault{\uhorn}{T5}

\usepackage{amsmath}
\usepackage{amsfonts}       
\usepackage{amssymb}
\usepackage{amsthm}
\usepackage{bbm}
\usepackage{enumerate}
\usepackage{enumitem,kantlipsum}

\usepackage{cleveref}
\usepackage{mathtools}
\usepackage{xspace}
\usepackage[disable]{todonotes}
\usepackage{caption,subcaption}
\usepackage{multirow}
\usepackage{subcaption}

\usepackage{comment}
\usepackage{booktabs}
\usepackage{tikz}
\usepackage{ltablex}
\usetikzlibrary{bayesnet,calc,matrix,positioning,shapes,backgrounds}

\definecolor{lightgray}{gray}{0.85}
\definecolor{lightlightgray}{gray}{0.9}
\definecolor{C1}{HTML}{1F77B4}
\definecolor{C2}{HTML}{FF7F0E}
\definecolor{C3}{HTML}{2CA02C}
\definecolor{C4}{HTML}{D62728}
\definecolor{C5}{HTML}{9467BD}

\crefname{section}{\S}{\S\S}
\Crefname{section}{\S}{\S\S}
\crefname{table}{Tab.}{}
\crefname{figure}{Fig.}{}
\crefname{algorithm}{Algorithm}{}
\crefname{equation}{Eq.}{}
\crefname{appendix}{App.}{}
\crefname{thm}{Theorem}{}
\crefname{prop}{Proposition}{}
\crefname{cor}{Corollary}{}
\crefname{observation}{Observation}{}
\crefname{assumption}{Assumption}{}
\crefformat{section}{\S#2#1#3}

\usepackage{tikz}

\usepackage{bbm}
\makeatletter
\newcommand*\iftodonotes{\if@todonotes@disabled\expandafter\@secondoftwo\else\expandafter\@firstoftwo\fi}  
\makeatother
\newcommand{\noindentaftertodo}{\iftodonotes{\noindent}{}}
\newcommand{\note}[4][]{\todo[author=#2,color=#3,size=\scriptsize,fancyline,caption={},#1]{#4}} 

\definecolor{dandelion}{HTML}{FFD464}
\newcommand{\lucas}[2][]{\note[#1]{lucas}{dandelion!60}{#2}}

\newcommand{\ryan}[2][]{\note[#1]{ryan}{violet!40}{#2}}

\newcommand{\karolina}[2][]{\note[#1]{karolina}{orange!10}{#2}}

\newcommand{\edo}[2][]{\note[#1]{edo}{green!10}{#2}}
\newcommand{\Edo}[2][]{\edo[inline,#1]{#2}\noindentaftertodo}

\DeclareMathOperator*{\argmax}{argmax}

\newcommand{\R}{\mathbb{R}}

\def\calD{{\mathcal{D}}}

\def\vh{{\boldsymbol{h}}}

\def\vtheta{{\boldsymbol{\theta}\xspace}}

\def\vphi{{\boldsymbol{\phi}\xspace}}

\let\originalleft\left
\let\originalright\right
\renewcommand{\left}{\mathopen{}\mathclose\bgroup\originalleft}
\renewcommand{\right}{\aftergroup\egroup\originalright}

\newcommand{\sqr}[1]{\left[#1\right]}

\newcommand{\expectq}{\mathop{\mathbb{E}}_{C \sim \qphi}}

\newcommand{\bert}{BERT\xspace}
\newcommand{\mbert}{m-\bert}
\newcommand{\xlmr}{XLM-R\xspace}
\newcommand{\xlmrbase}{\xlmr-base\xspace}
\newcommand{\xlmrlarge}{\xlmr-large\xspace}

\newcommand{\ptheta}{p_{\vtheta}}

\newcommand{\qphi}{q_{\vphi}}

\newcommand{\prop}[1]{\textsc{#1}}

\newcommand{\NLL}{\mathcal{L}}
\newcommand{\entropy}{\mathrm{H}}

\newcommand{\XX}{{43}\xspace}

\newcommand*{\nameadjunct}{\relax}
\makeatletter
\renewcommand*{\NAT@nmfmt}[1]{\NAT@up #1\nameadjunct}
\makeatother

\newcommand*{\citeposs}[2][]{%
  \begingroup
  \renewcommand*{\nameadjunct}{'s}%
  \citet[#1]{#2}%
  \endgroup
}